\pdfoutput=1

\documentclass{llncs}
\usepackage{graphicx}
\usepackage{hyperref}
\usepackage{amssymb}
\usepackage{algorithm,algcompatible}
\usepackage{bbm}
\begin{document}
\mainmatter              
\title{Channel masking for multivariate time series shapelets}

\author{Dripta S. Raychaudhuri, Josif Grabocka \and Lars Schmidt-Thieme}
\authorrunning{Dripta S. Roychaudhuri et al.} 
\institute{Universit{\"a}t Hildesheim,
Information Sciences and Machine Learning Lab, Samelsonplatz 1,\\
31141 Hildesheim, Germany,\\
\email{dripta@ismll.de}}

\maketitle              

\begin{abstract}
Time series shapelets are discriminative sub-sequences and their similarity to time series can be used for time series classification. Initial shapelet extraction algorithms searched shapelets by complete enumeration of all possible data sub-sequences. Research on shapelets for univariate time series proposed a mechanism called shapelet learning which parameterizes the shapelets and learns them jointly with a prediction model in an optimization procedure. Trivial extension of this method to multivariate time series does not yield very good results due to the presence of noisy channels which lead to overfitting. In this paper we propose a shapelet learning scheme for multivariate time series in which we introduce channel masks to discount noisy channels and serve as an implicit regularization. 
\end{abstract}

\section{Introduction}
Time series classification is relevant in many different domains, such as health informatics \cite{11}, finance \cite{12} and bioinformatics \cite{14}. Time series are sequences of data points that have a temporal relation between each other. 

Originally, nearest neighbour models with Euclidean and later Dynamic Time Warping distance measures have been used for time series classification. Recent research on this topic focuses on supervised feature extraction where the goal is to identify motifs or local patterns that have discrimination qualities towards the target variable. One popular method is to identify shapelets \cite{1}. Shapelets are discriminative sub-sequences where the distance between a shapelet and its best matching sub-sequence on a time series is a good predictor for time series classification. Many methods try to find shapelets and apply Shapelet Transformation \cite{2}, which converts the raw time series data using the shapelets to a different representation that containing features that correspond to a specific shapelet where its value is the minimal distance to the time series. 

Grabocka et al. \cite{3} proposed a novel approach that learns shapelets, by optimizing a classification loss as objective function, known as the LTS approach. They learned shapelets whose minimal distances to time series instances can be used as features for a logistic regression model.

However, learning shapelets across all channels for multivariate time series is not optimal as noisy channels affect the prediction in a strong way. This paper aims at discounting the noisy channels, while promoting the informative ones. To achieve this we introduce channel masks for each series channel. The dimension wise weighted impact is learned automatically along with the shapelets by numerical optimization of the loss.

The proposed approach was evaluated on 9 real world data sets
from the UCR \cite{10} and UCI \cite{15} repositories as well as a synthetic data set. The experimental results show that the proposed approach outperforms the existing methods on majority of data sets.

\section{Related Work}
Most existing time series classification methods fall into two categories, distance based methods and feature based methods \cite{9}. 

Distance based methods measure the similarity between two time series using a similarity metric and consequently, classification can be done using algorithms such as k-Nearest Neighbour or Support Vector Machines. For feature based methods, each time series is characterized by a feature vector and any feature classifier can be used to produce classification. 

The concept of shapelets, discriminative sub-sequences, was first introduced by Ye et al. \cite{1}. The idea relies on considering all sub-sequences of the training data and assess them regarding a scoring function to estimate how predictive they are with respect to the target. Since a time series dataset usually contains many shapelet candidates, a brute-force search is time-consuming. Grabocka et al. \cite{3} suggest learning optimal shapelets with respect to the target and report statistically significant improvements in accuracy compared to other shapelet-based classifiers.

A common approach for multivariate time series classification is to apply dimensionality reduction like singular value decomposition on the data and then use any classifier on the new projected data. This overcomes the problem of time series having varying lengths \cite{4,5}. Other methods use similarity-based methods that have proven to be useful for
univariate time series classification. For example dynamic time warping was applied on multivariate time series in the context of accelerometer-based gesture recognition \cite{6,7}.

Currently, the state of the art in terms of accuracy is based on extracting a supervised symbolic representation from multivariate time series \cite{8}. Random forests are learned using time-series values, their derivatives, time indices and the frequencies of leaf classes from the learned random forest classifiers are used to yield a supervised bag-of-patterns representation. The accuracy of a second layer of random forest over the bag-of-symbolic patterns is shown to yield state-of-the  art results.

Furthermore, feature weighting is broadly used in batch learning \cite{16} to assign different weights to feature accordingly to their relevance to the concept to be learned and improve prediction accuracy. 

In contrast to the related work, we hypothesize that features can be weighted dynamically in order to augment the importance of relevant features and diminish the importance of those which are deemed irrelevant. 

\section{Proposed Method}

\subsection{Background}

\subsubsection{Time Series Dataset}
A multivariate time series is a set of ordered values arranged in temporal order. We assume that the dataset is composed of $I$ training instances, $V$ channels and for notation ease we assume that each series contains $Q$-many ordered values, even though our method can operate on variable series lengths. The series target is a nominal variable $Y \in \{$1,\dots,C$\}^{I}$ having $C$ categories.
\subsubsection{Shapelets}
Informally, shapelets are time series sub-sequences which are maximally representative of a class. We denote a shapelet by $P \in \mathbb{R}^{K\times L\times V}$ and the length of a shapelet by $L$, and the total number of shapelets to be learned by $K$. Each shapelet has, in accordance with the time series dataset, $V$ channels.
\subsubsection{Masks}
For quantifying the predictive power of a channel in the time series data, masks are learned for each shapelet. We will denote a mask by $\mu_{k,v}$ where $k$ indicates the shapelet with which the particular mask is associated and $v$ indicates the channel. 
\subsubsection{Calculating distance between shapelet and time series}
A sliding window strategy is used for calculating the distance between the i-th time series $T_i$ and the k-th shapelet $P_k$. All possible segments of $T_i$ can be extracted by sliding a window of size $L$ (which is also the length of shapelet) across all channels of $T_i$. Formally, the segment of time series $T_i$ starting at time-stamp $j$ is defined as $T_j$,....,$T_{j+L-1}$ and will be denoted by $T_{i,v,j}$, where $v$ represents a particular channel. If the starting index of the sliding window is incremented by one, then there are total $J = Q - L + 1$ segments.

The distance is averaged across all the channels and the masks are overlayed to get the actual distance $A_{i,k}$ as shown in Equation 1.

\begin{equation}
A_{i,k}
= \min_{j=1,....J} \frac{1}{V \times L}\sum_{v=1}^{V}  \mu_{k,v} \times \sum_{l=1}^{L}{(T_{i,v,j+l-1}-P_{k,v,l})}^2
\end{equation}

\subsection{Learning multivariate shapelets with channel masks}
\subsubsection{Model}
In this section we describe our model in detail with regards to the parameters. We show how the shapelets, masks and the weights of the logistic regression model are learned simultaneously through stochastic gradient descent.

The computed minimum distances from the shapelets act as features, which can then be fed into a linear classifier to predict the target variable,

\begin{equation}
Z_{i} = W_0 + \sum_{k=1}^{K}{A_{i,k}W_{k}}  \ \ \ \forall i \ \in \ \{1, 2,...., I\}
\end{equation}
\begin{equation}
\hat{Y_{i}} = \sigma (Z_{i}) \ \ \  \forall i \ \in \ \{1, 2,...., I\}
\end{equation}
where $\sigma(.)$ represents the sigmoid function.

Our model learns $K$ many shapelets of varying lenghts. The length of a shapelet is randomly chosen starting at a minimum $L^{min}$ and restricted by a maximum value $L^{max}$. The length of a shapelet $k$ is denoted by $L_k = L^{min} + r \times L^{max} \ \ \ \forall k \in \{1,2,\dots,K\}$, where $r$ is a random number uniformly distributed between 0 and 1. The shapelets, therefore, can be defined as $P \in  \mathbb{R}^{K\times V\times *} $.

In order to learn from multi-class targets $Y \in \{1,2,\dots,C\}^I$ with C categories, we will convert the problem into C-many one-vs-all sub-problems and train C classifiers. The one-vs-all binary targets are defined in Equation 4.
\begin{equation}
Y_{i,c} = \left\{
        \begin{array}{ll}
            1, \quad Y_i=c\\   
            0, \quad Y_i\neq c
       \end{array}
       \ \  \forall i \ \in \ \{1, 2,\dots, I\} \ \forall c \ \in \ \{1, 2,\dots, C\}\\ 
    \right.
\end{equation}
Now, the target prediction is modified for generalization towards the multi-class case. This is shown below.

\begin{equation}
Z_{i,c} = W_{0,c} + \sum_{k=1}^{K}{A_{i,k}W_{k,c}} 
\end{equation}

\begin{equation}
\hat{Y}_{i,c} = \frac{\exp(Z_{i,c})}{\sum_{c=1}^{C}\exp(Z_{i,c})}
\end{equation}

We train our model by minimizing the cross-entropy or the logistic loss function between the true targets $Y$ and the predicted values $\hat{Y}$.

\begin{equation}
L(Y,\hat{Y}) = -Ylog(\hat{Y}) -(1-Y)log(1-\hat{Y})
\end{equation}

The logistic loss function, along with the $L2$ regularization term represent the regularized objective function, denoted as $F$ in Equation 8. Our objective is to jointly learn the shapelets $P$, channel masks $\mu$ and the optimal hyperplane $W$ that minimize the classification objective.

\begin{equation}
{\arg\min}_{P,\mu,W} F = \sum_{i=1}^{I}\sum_{c=1}^{C}L(Y_i,\hat{Y}_i) + \frac{\lambda_W}{2}\|W\|^2
\end{equation}

\subsubsection{Activation function on masks}
The masks of series channels can take both positive and negative values. However, the concept of negative values of masks make less sense, so we decide to apply an activation function on the masks to make them positive. The ideal activation function for this purpose should always output positive real numbers and be monotonically increasing.

We considered two such functions: the sigmoid function and the rectified linear function. In our experiments, the rectified linear function offered better performance compared to the sigmoid. Thus, the former was selected as the activation function of choice over the masks. Further information regarding the choice of the activation function can be found in the section 4. 

Consequently, Equation 1 is modified as,

\begin{equation}
A_{i,k}
= \min_{j=1,....J} \frac{1}{V \times L}\sum_{v=1}^{V}  f(\mu_{k,v}) \times \sum_{l=1}^{L}{(T_{i,v,j+l-1}-P_{k,v,l})}^2
\end{equation}

\begin{equation}
f(x) = \max(0,x)
\end{equation}
where $f(x)$ is the rectified linear function.

\subsubsection{Decomposed objective}
The optimization procedure we will adopt in our model is a stochastic gradient descent approach that ameliorates the error caused
by one instance of the training set at a time. Equation 11 demonstrates the decomposed objective function $F_i$, which corresponds to a division of the objective of Equation 8 into per-instance, per-class, losses for each time series.

\begin{equation}
F_{i,c} = L(Y_i,\hat{Y}_i) + \frac{\lambda_W}{2IC}\sum_{k=1}^{K}\sum_{c=1}^{C}{W_{k,c}}^2
\end{equation}

\subsubsection{Gradients for shapelets}
The gradient of point $\ell$ in shapelet $k$ with respect to the objective of the $i$-th time series is defined in Equation 12.

\begin{equation}
\frac{\partial F_{i,c}}{\partial P_{k,v,l}} = \frac{\partial L(Y_{i,c},\hat{Y}_{i,c})}{\partial Z_{i,c}}\frac{\partial Z_{i,c} }{\partial A_{i,k}}\frac{\partial A_{i,k}}{\partial P_{k,v,l}}
\end{equation}

Also, the gradient of the loss with respect to the predicted
target and the gradient of the predicted target with respect
to the minimum distances is shown in Equation 13.

\begin{equation}
\frac{\partial L(Y_{i,c},\hat{Y}_{i,c})}{\partial Z_{i,c}} = \hat{Y}_{i,c} - Y_{i,c}
\ \  \ \frac{\partial Z_{i,c}}{A_{i,k}} = W_{k,c}
\end{equation}

Finally, we compute the gradient of the overall minimum distance with respect to the shapelets.
\begin{equation}
\frac{\partial A_{i,k}}{\partial P_{k,v,l}} = \frac{-2}{V \times L} f(\mu_{k,v}){(T_{i,v,j^*+l-1}-P_{k,v,l})}
\end{equation}
The index $j^*$ in Equation 14 indicates the starting time index in a time series where the distance from the shapelet to that time series is minimal.
\subsubsection{Gradients for masks}
The channel masks are also learned via gradient descent to discount the inconsequential channels while promoting the essential ones at the same time. The gradient depends intrinsically on the choice of the activation function.
\begin{equation}
\frac{\partial F_{i,c}}{\partial \mu_{k,v}} = \frac{\partial L(Y_{i,c},\hat{Y}_{i,c})}{\partial Z_{i,c}}\frac{\partial Z_{i,c} }{\partial A_{i,k}}\sum_{l=1}^{L}\frac{\partial A_{i,k}}{\partial \mu_{k,v}}
\end{equation}
\begin{equation}
\frac{\partial F_{i,c}}{\partial \mu_{k,v}} = (\hat{Y}_{i,c} - Y_{i,c})W_{k,c}\sum_{l=1}^{L}\frac{1}{V \times L} f'(\mu_{k,v}){(T_{i,v,j^*+l-1}-P_{k,v,l})^2}
\end{equation}
If the activation function is chosen as sigmoid, the derivative is defined as $f'(x) = f(x)(1-f(x))$ while the derivative for the rectified linear activation is $f'(x) = \mathbbm{1}_\{x>0\}$.

\subsubsection{Gradients for weights of classifier}
The hyper-plane weights W are also learned to minimize
the classification objective via stochastic gradient descent.
The partial gradient for updating each weight $W_{k,c}$ and the bias terms $W_{0,c}$, is defined in Equations 17.
\begin{equation}
\frac{\partial F_{i,c}}{\partial W_{k,c}} = (\hat{Y}_{i,c} - Y_{i,c})A_{i,k} + \frac{\lambda_W}{IC}W_{k,c}
\ \ \ \ \ \frac{\partial F_{i,c}}{\partial W_{0,c}} = (\hat{Y}_{i,c} - Y_{i,c})
\end{equation}

\subsubsection{Learning algorithm}
We have derived the gradients of the shapelets, masks and the weights, so we now present the overall learning algorithm. The steps of the learning process are shown in Algorithm 1. Instead of the vanilla stochastic gradient descent, we use the AdaGrad optimization procedure in our learning algorithm. Shapelets, masks and weights are all  initialized by small random values drawn from a normal distribution with mean 0 and variance 1. The convergence of Algorithm 1 depends on two parameters, the
learning rate $\eta$ and the maximum number of iterations. 

The learning rate and the number of iterations are learned via cross-validation from the training data.

\setlength{\textfloatsep}{0pt}
\begin{algorithm}
\caption{Learning channel masks and shapelets}\label{alg:euclid}
\textbf{Input}: Time series $T \in \mathbb{R}^{I\times V\times Q}$, Binary labels $Y \in \mathbb{R}^{I\times C}$, Number of Shapelets $K$, Minimum Shapelet Length $L_{min}$, Minimum Shapelet Length $L_{max}$, Regularization $\lambda_W$, Learning Rate $\eta$, Number of iterations: $maxIter$ \\
\textbf{Output}: Shapelets $P \in \mathbb{R}^{K\times V\times *}$, Masks $\mu \in \mathbb{R}^{K\times V}$, Classification weights $W \in \mathbb{R}^{K\times C}$, $W_0 \in \mathbb{R}^{C}$
\begin{algorithmic}[1]
\STATE Initialize $P$, $\mu$, $W$, $W_0$
\FOR{$iter=1$ to $maxIter$}

    \FOR{$i=1$ to $I$}
        \FOR{$k=1$ to $K$}
            \STATE $A_{i,k} = \min_{j=1,....J} \frac{1}{V \times L}\sum_{v=1}^{V}  f(\mu_{k,v}) \times \sum_{l=1}^{L}{(T_{i,v,j+l-1}-P_{k,v,l})}^2$
        \ENDFOR
        
        \FOR{$c=1$ to $C$}
            \STATE $Z_{i,c} = W_{0,c} + \sum_{k=1}^{K}{A_{i,k}W_{k,c}} $
        \ENDFOR
        
        \FOR{$c=1$ to $C$}
            \STATE $\hat{Y}_{i,c} = \frac{\exp(Z_{i,c})}{\sum_{c=1}^{C}\exp(Z_{i,c})}$
        \ENDFOR
        \STATE Update $\eta_{P_{k,v,l}}, \mu_{P_{k,v}}, \eta_{W_{k,c}},     \eta_{W_{0,c}}$ via AdaGrad
        \FOR{$c=1$ to $C$}
            \FOR{$k=1$ to $K$}
                \FOR{$v=1$ to $V$}
                    \FOR{$l=1$ to $L$}
                        \STATE$P_{k,v,l} = P_{k,v,l} - \eta_{P_{k,v,l}}\frac{\partial F_{i,c}}{\partial P_{k,v,l}}$
                        \STATE$\mu_{k,v} = \mu_{k,v} - \mu_{P_{k,v}}\frac{\partial F_{i,c}}{\partial \mu_{k,v}}$
                    \ENDFOR
                \ENDFOR
                \STATE$W_{k,c} = W_{k,c} - \eta_{W_{k,c}}\frac{\partial F_{i,c}}{\partial W_{k,c}}$
            \ENDFOR
            \STATE$W_{0,c} = W_{0,c} - \eta_{W_{0,c}}\frac{\partial F_{i,c}}{\partial W_{0,c}}$
        \ENDFOR
    \ENDFOR
\ENDFOR
\STATE \textbf{return} $P$, $\mu$, $W$, $W_0$
\end{algorithmic}
\end{algorithm}
\setlength{\dbltextfloatsep}{0pt}
\section{Experiments}

We evaluated our method on 9 real world data sets using the default train and test splits and on a synthetic data set created by us. Table 2 summarizes the details about all the data sets. In the next section, we explain the creation of the synthetic data set.

\subsection{Experiments on synthetic dataset}

Consider a multivariate time series classification task where the class of a particular time series depends on patterns appearing in the first couple of channels, and the rest of the channels contain noise or contain patterns which do not contribute to the classification.
Methods which learn to recognize patterns across all the channels of the time series will perform poorly in such cases due to over-fitting. Thus, we need a channel selection procedure to highlight the channels with high predictive power.

The idea is to plant patterns across channels, however, to highlight the channel weighting property of our method, we plant patterns in a principled manner.

\begin{figure}
\centering
  \includegraphics[width=0.6\linewidth]{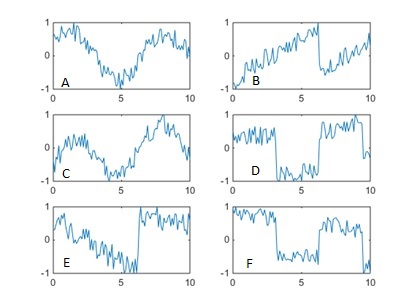}
  \caption[]{Patterns implanted in the synthetic dataset.}
  \label{fig:patterns}
\end{figure}

We construct a synthetic data set with three classes $C_1$, $C_2$ and $C_3$ and 40 channels. The first step involves generating patterns. For our purpose, we create 6 patterns, denoted as $A,B,\dots,F$ which are shown in Figure 2. In the next step, we sample the class from a uniform distribution, i.e. $y\sim p(Y)$. Following this, we plant patterns in the first two channels conditioned on the class, according to the distribution shown below.
\\
 \begin{table}
 \centering
 \begin{tabular}{|c|c|c|c|c|} 
 \hline
 $X_1$ & $X_2$ & $ p(X_1,X_2|Y=1)$ & $ p(X_1,X_2|Y=2)$ & $ p(X_1,X_2|Y=3)$ \\ \hline
 A & D & 0.5 & 0.0 & 0.0 \\ 
 \hline
 B & F & 0.5 & 0.0 & 0.0 \\
 \hline
 B & E & 0.0 & 0.5 & 0.0 \\
 \hline
 C & D & 0.0 & 0.5 & 0.0 \\
 \hline
 C & F & 0.0 & 0.0 & 0.5 \\ 
 \hline
 A & E & 0.0 & 0.0 & 0.5 \\
 \hline
\end{tabular}
\caption[]{Likelihood of pattern presence in the first two channels}
\end{table}

The remaining channels are filled by random patterns according to a uniform distribution, independent of the class, i.e., $x_c\sim p(X_C ) \ \ \forall C \in \{$3,4,5,\dots,40$\}$. The rationale behind creating the dataset in this manner is to endow the first two channels with predictive power, while the rest of the channels contain random patterns that confuse methods that try to learn patterns across all channels. A trivial dataset creation involved filling rest of the channels with noise, but this would be partially overcome by patterns learning to ignore the noisy channels themselves. 

\subsubsection{Choice of activation function}
We compare two functions which act on the raw values of the masks and produce non-negative masking weights: the sigmoid and the rectified linear unit. According to our experiment, the rectifed linear function gives better accuracy.

\begin{figure}[htp]

\centering
\includegraphics[width=.45\textwidth]{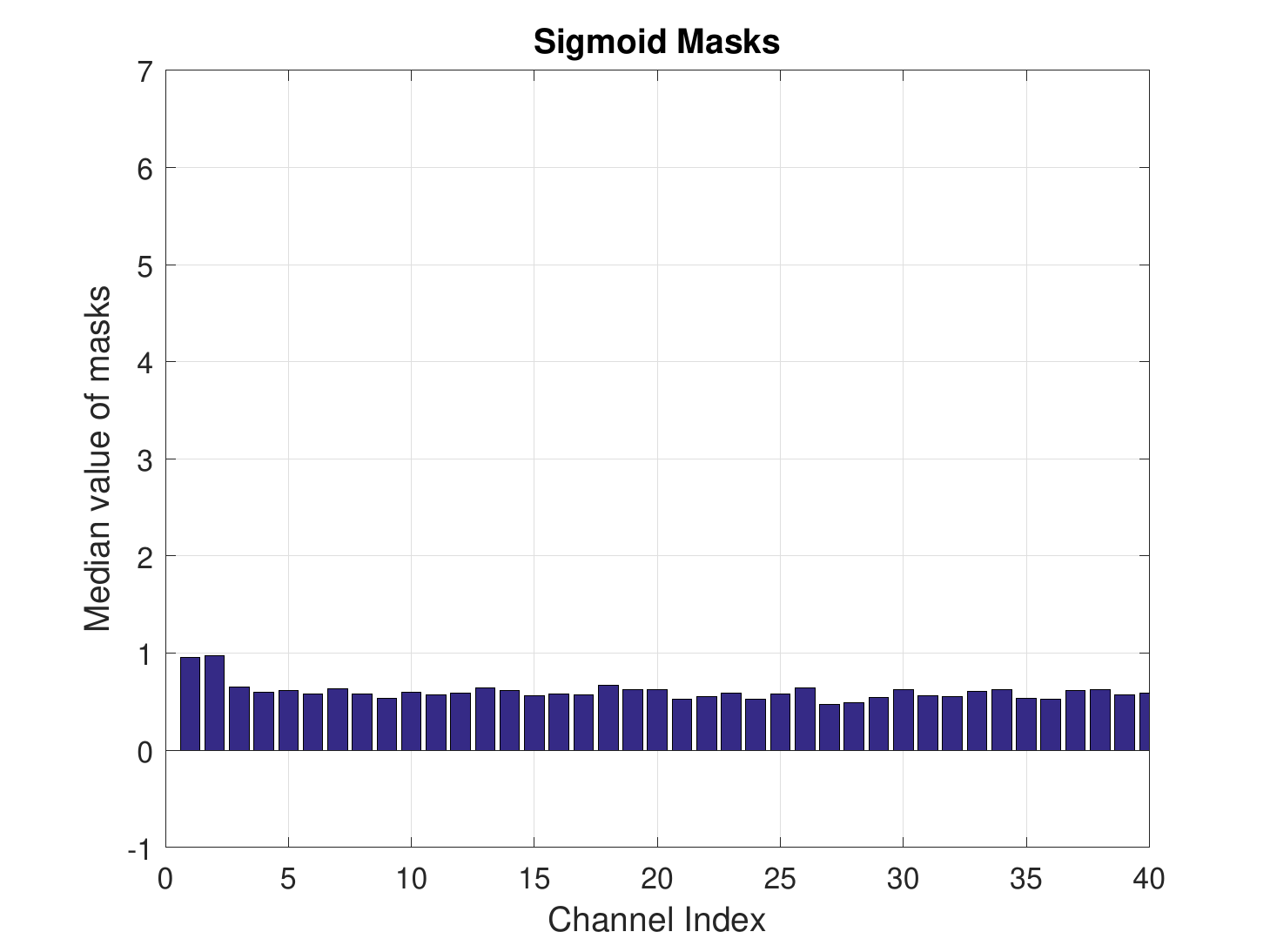}\hfill
\includegraphics[width=.45\textwidth]{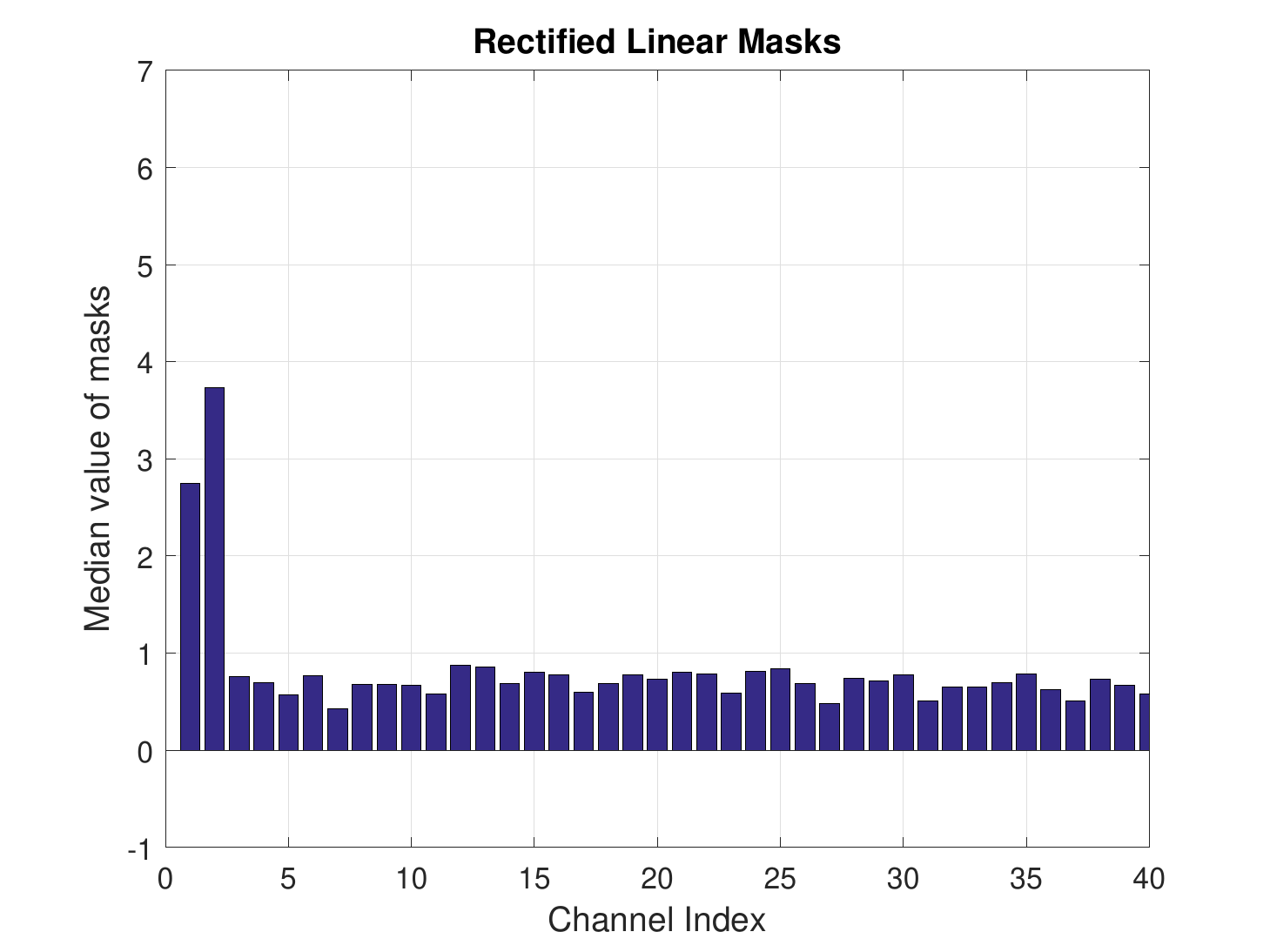} 
\caption[]{Comparing different activation functions}
\end{figure}

The problem of using the sigmoid function is that it squashes every weight value between 0 and 1, and the significance of the first two channels is negligible, as evident from the first figure. On the other hand, the rectified linear function allows the mask weights to grow beyond 1 and the difference between the weights of the first two channels and the biggest weight of the other channels is much more significant.

\subsubsection{Evolution of masks on the synthetic dataset}
\begin{figure}[htp]

\centering
\includegraphics[width=.3\textwidth]{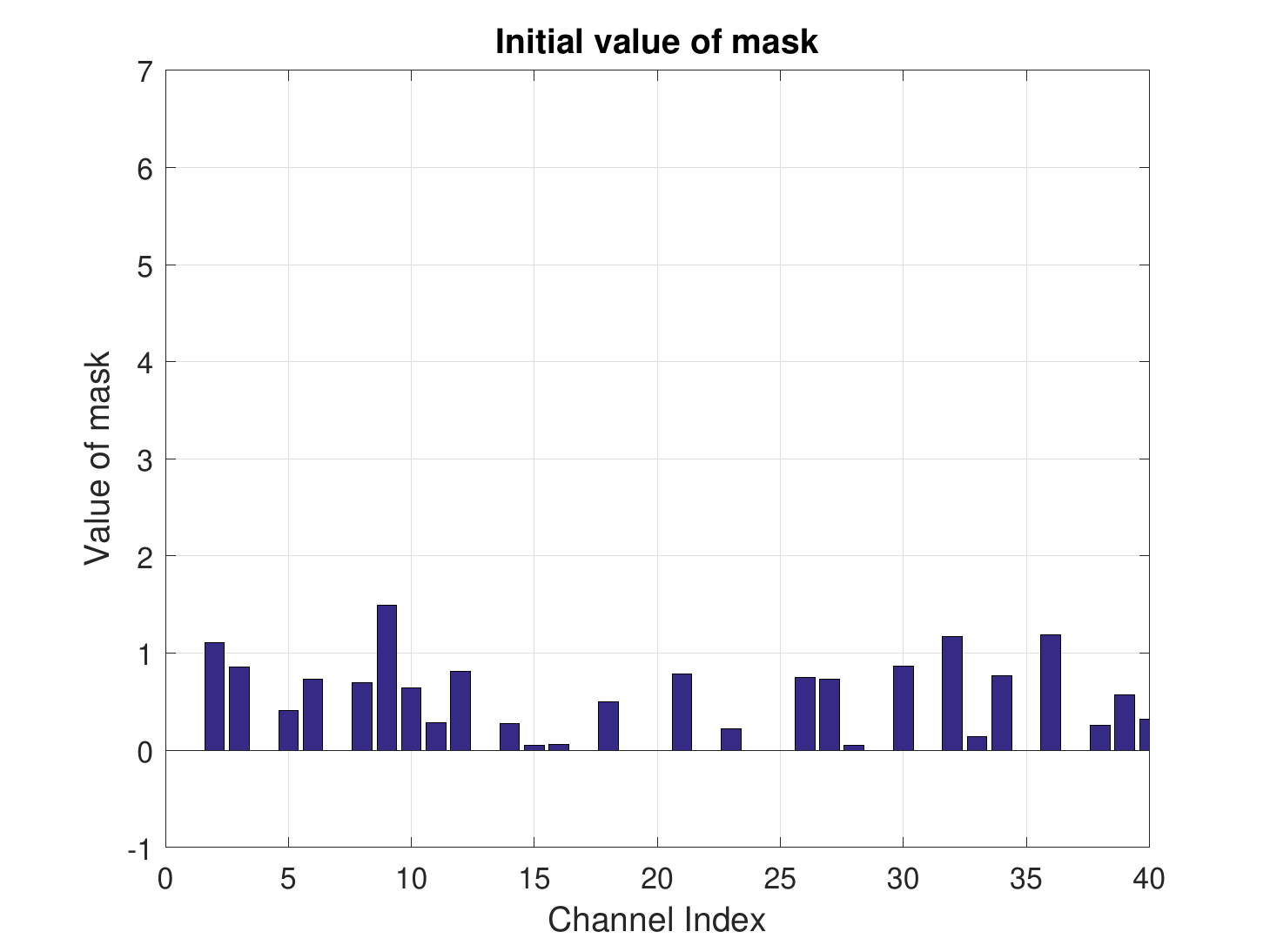}\hfill
\includegraphics[width=.3\textwidth]{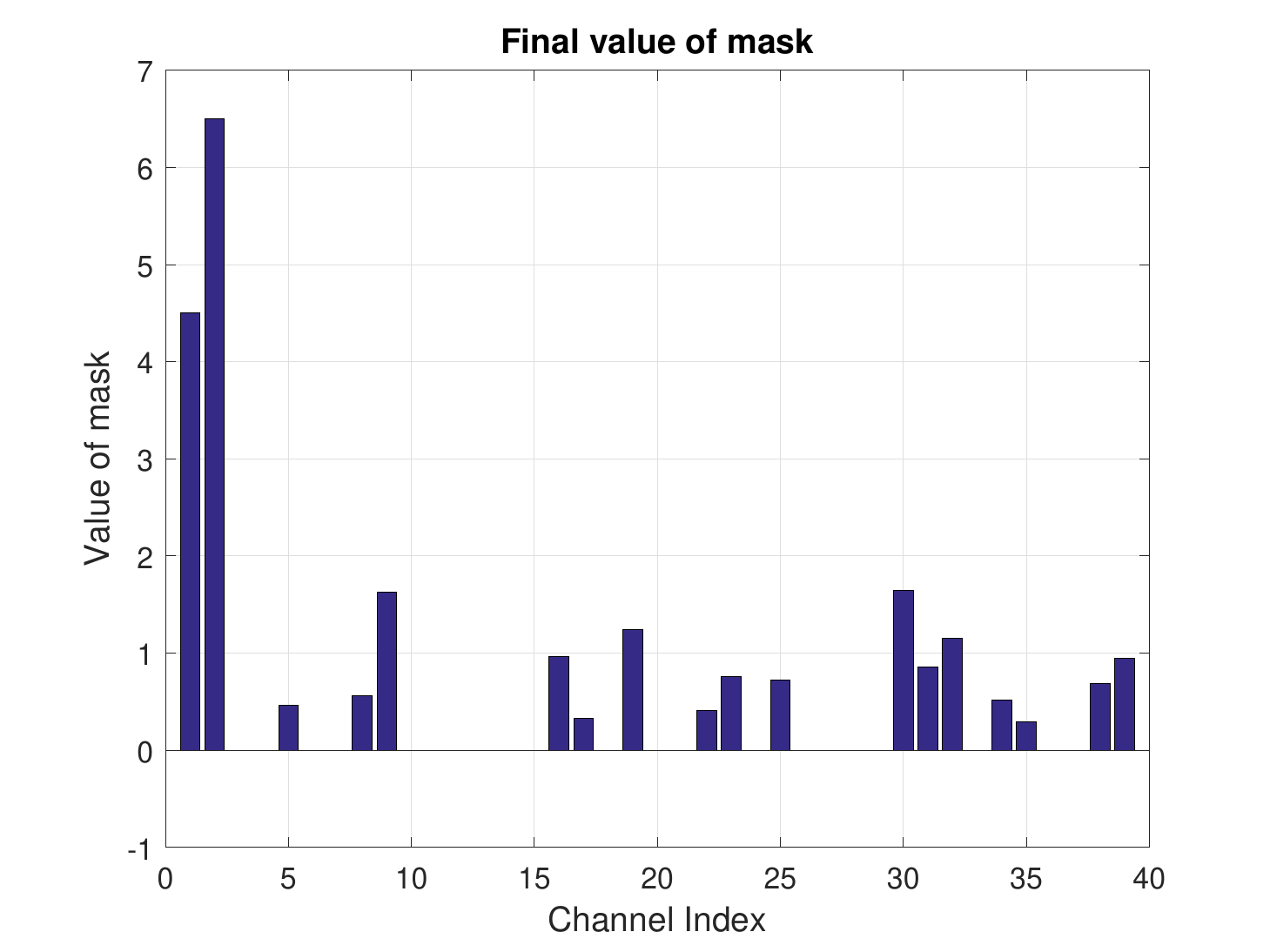}\hfill
\includegraphics[width=.3\textwidth]{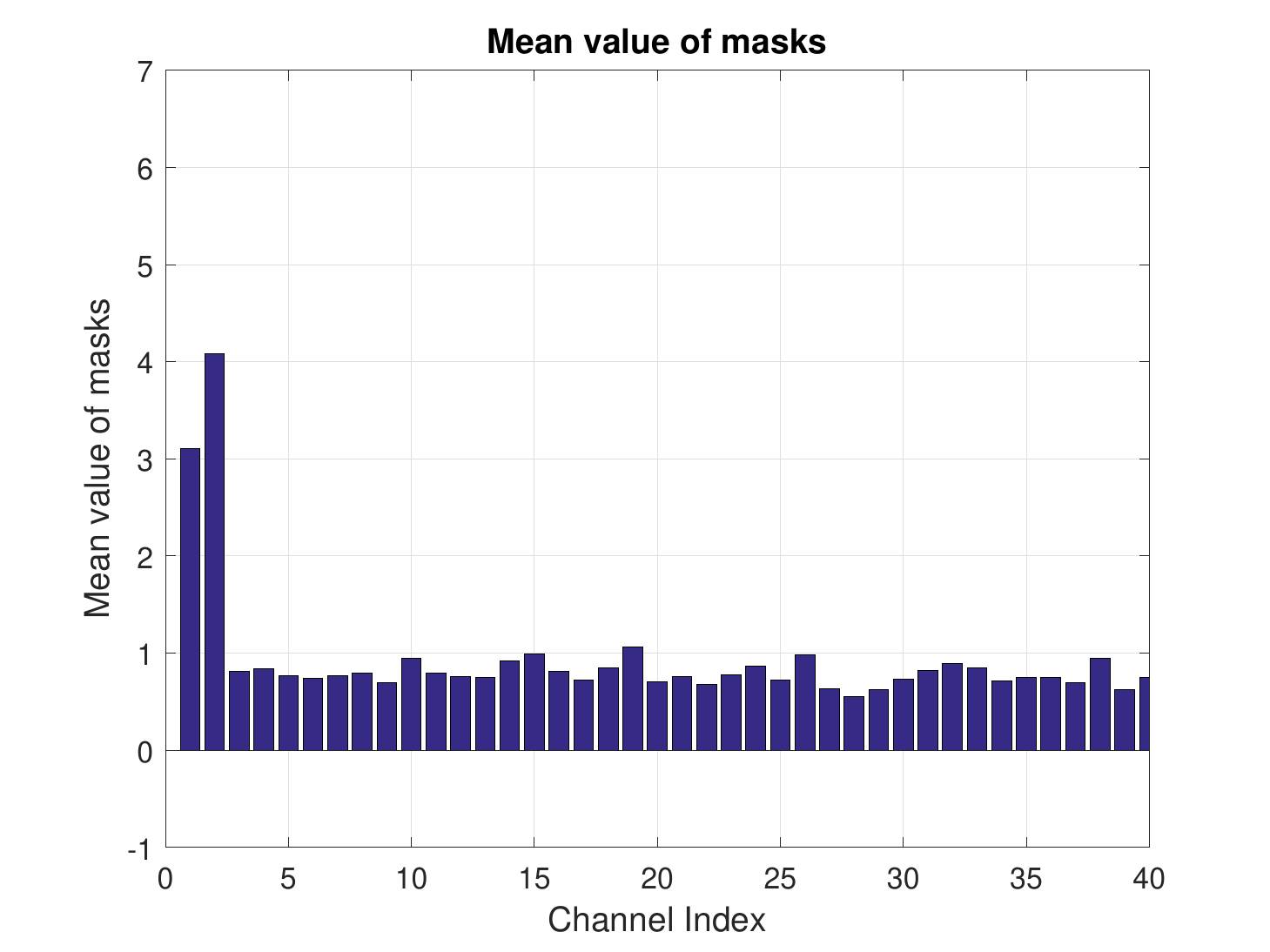}
\caption[]{Evolution of a mask on the synthetic dataset}
\end{figure}
In Figure 3 we illustrate the evolution of a mask from initialization to just after 100 iterations. As expected, the first two channels receive the highest weights as the patterns contained in these channels directly determine the class of the instance. 


\subsection{Hyper-parameter search}
Hyper-parameters were found through a grid search approach using five fold cross-validation over the training data. The
number of shapelets was searched in a range of $K \in \{10,20,40,100\}$. The regularization parameter was one
of $\lambda_W \in \{0.001,0.01,0.1\}$. The learning rate was kept fixed at
a small value of $\eta=0.1$, while the number of iterations was fixed at 1000.
\begin{table}
 \centering
 \begin{tabular}{|c|c|c|c|c|c|} 
 \hline
 Dataset & Train/Test instances & Channels & Length & Classes & $\lambda_W/K$ \\ \hline
 Synthetic&500/200&40&202&2&0.01/20 \\ 
 \hline
 Wafer&298/896&6&104-198&2&0.01/100 \\ 
 \hline
 CMUsub16&29/29&62&127-580&2&0.1/10 \\
 \hline
 ECG&100/100&2&39-152&2&0.1/100 \\
 \hline
 JapaneseVowels&270/370&12&7-29&9&0.001/100  \\
 \hline
 PenDigits&300/10692&2&8&10&0.1/100 \\ 
 \hline
 ArabicDigits&6600/2200&13&4-93&10&0.1/100 \\
 \hline
 LIBRAS&180/180&2&45&15&0.01/100 \\
 \hline
 CharacterTrajectories&300/2558&3&109-205&20&0.01/100 \\
 \hline
 uWave&200/4278&3&315&8&0.01/100 \\
 \hline
\end{tabular}
\caption[]{Dataset information and parameter search results}
\end{table}

\subsection{Accuracy}
The 9 multivariate datasets used to evaluate our method are from various domains such as handwriting recognition (CharacterTrajectories,
PenDigits), motion (CMU MOCAP S16), gesture (uWaveGestureLibrary)
and speech recognition (ArabicDigits, JapaneseVowels). Detailed characteristics of the datasets are given in Table 2 and detailed results are presented in Table 3. We compare our method to Symbolic Representation for Multivariate Timeseries (SMTS) \cite{8}, nearest neighbor with dynamic time warping distance without a warping window (NNDTW) and a multivariate extension of TSBF (MTSBF) \cite{11}. We also list the results of the shapelet method without masks to reinforce the efficacy of our method. It is clear from the results that masking improves accuracy.

\begin{table}
 \centering
 \begin{tabular}{|c|c|c|c|c|c|} 
 \hline
 $Dataset$ & $Masked Shapelet$ & $Shapelet$ & $SMTS$ & $ NN-DTW$ & $MTSBF$ \\ \hline
 Synthetic&\textbf{0.000}&0.195&-&-&- \\ 
 \hline
 Wafer&\textbf{0.014}&0.019&0.035&0.023&0.015 \\ 
 \hline
 CMUsub16&\textbf{0.000}&\textbf{0.000}&0.003&0.069&0.003 \\
 \hline
 ECG&\textbf{0.110}&0.170&0.182&0.150&0.165 \\
 \hline
 JapaneseVowels&\textbf{0.019}&0.027&0.031&0.351&-  \\
 \hline
 PenDigits&\textbf{0.053}&0.102&0.083&0.088&- \\ 
 \hline
 ArabicDigits&\textbf{0.036}&0.065&\textbf{0.036}&0.092&- \\
 \hline
 LIBRAS&0.111&0.177&\textbf{0.091}&0.200&- \\
 \hline
 CharacterTrajectories&0.018&0.025&\textbf{0.008}&0.040&0.033 \\
 \hline
 uWave&0.071&0.076&\textbf{0.059}&0.071&0.100 \\
 \hline
\end{tabular}
\caption[]{Accuracy for the real world datatsets}
\end{table}

It is very evident that channel masking improves a trivial extension of the learning shapelets method to multivariate time series. In all the datasets tested masked shapelets outperforms the baseline. Compared to the state-of-the-art SMTS method, masked shapelets performs better, giving lower error rates on 6 out of the 9 data sets tested.  

\section{Conclusion}
In this paper, we have non-trivially extended the Learning Time-series Shapelets method to multivariate time series classification by using the novel concept of channel masking. It outperformed state-of-the-art methods in multivariate time series classification. Future work in multivariate time series classification can benefit through this concept of channel masking.


\begin{thebibliography}{20}

\bibitem[1]{1}
L. Ye, E. Keogh: 
Time Series Shapelets: A New Primitive for Data Mining. 
Proceedings of the 15th ACM SIGKDD International Conference on Knowledge Discovery and Data Mining, KDD '09, ACM, New York, NY, USA, 2009, pp. 947-–956.

\bibitem[2]{2}
J. Hills, J. Lines, E. Baranauskas, J. Mapp, A. Bagnall: 
Classification of time series by shapelet transformation. 
Data Mining and Knowledge Discovery 28 (4) (2014) 851-–881.


\bibitem[3]{3}
J. Grabocka, N. Schilling, M. Wistuba, L. Schmidt-Thieme: 
Learning Time-Series Shapelets. 
Proceedings of the 20th ACM SIGKDD International Conference on Knowledge Discovery and Data Mining, ACM, 2014.

\bibitem[4]{4}
C. Li, L. Khan, B. Prabhakaran: 
Real-time classification of variable length multi-attribute motions.
Knowl. Inf. Syst. 10 (2) (2006) 163–-183.

\bibitem[5]{5}
X. Weng, J. Shen: 
Classification of multivariate time series using locality preserving projections.
Knowledge-Based Systems 21 (7) (2008) 581–-587.

\bibitem[6]{6}
A. Akl, S. Valaee: 
Accelerometer-based gesture recognition via dynamic-time warping, affinity propagation and compressive sensing. 
ICASSP’10, 2010, pp. 2270-–2273.

\bibitem[7]{7}
J. Liu, L. Zhong, J. Wickramasuriya, V. Vasudevan: 
uWave: Accelerometer-based Personalized Gesture Recognition and Its Applications. 
Pervasive Mob. Comput. 5 (6) (2009) 657-–675.

\bibitem[8]{8}
M. Baydogan, G. Runger:
Learning a symbolic representation for multivariate time series classification. 
Data Mining and Knowledge Discovery (2014) 1-–23.

\bibitem[9]{9}
Z.Xing, J.Pei, E.Keogh: 
A brief survey of sequence classification. 
ACM SIGKDD Explorations Newsletter, 12(1):40--48, 2010.

\bibitem[10]{10}
E. Keogh: 
The UCR time series classification/clustering homepage: \url{http://www.cs.ucr.edu/~eamonn/time\_series\_data/} (2014).

\bibitem[11]{11}
M. Baydogan, G. Runger, E. Tuv: 
A Bag-of-Features Framework to Classify Time Series. 
IEEE Transactions on Pattern Analysis and Machine Intelligence, 35 (11) (2013) 2796-–2802.

\bibitem[12]{12}
D.A. Clifton, K.E. Niehaus, P .Charlton, G.W. Colopy:
Health Informatics via Machine Learning for the Clinical Management of Patients.
Yearbook of Medical Informatics. 2015;10(1):38-43. 

\bibitem[13]{13}
C.W.J. Granger:
Forecasting stock market prices: Lessons for forecasters.
International Journal of Forecasting 8(1), 3–13 (1992).

\bibitem[14]{14}
R. Amato, A. Ciaramella, N. Deniskina, C. Del Mondo, D. di Bernardo, C. Donalek, G. Longo, G. Mangano, G. Miele, G. Raiconi, A. Staiano, R. Tagliaferri:
A multi-step approach to time series analysis and gene expression clustering. 
Bioinformatics, Volume 22, Issue 5, 1 March 2006, Pages 589–596.

\bibitem[15]{15}
M. Lichman: 
The UCI Machine Learning Repository \url{http://archive.ics.uci.edu/ml} 
Irvine, CA: University of California, School of Information and Computer Science.

\bibitem[16]{16}
N. Jankowski, K. Usowicz:
Neural Information Processing: 18th International Conference, ICONIP 2011, Shanghai, China, November 13-17, 2011,
Proceedings, Part II, chapter Analysis of Feature Weighting Methods Based on Feature Ranking Methods for Classification, pages 238-247. Springer Berlin Heidelberg, Berlin, Heidelberg, 2011.

\bibitem[17]{17}
J.P. Barddal, H. Murilo Gomes, F. Enembreck , B. Pfahringer, A. Bifet:
On Dynamic Feature Weighting for Feature Drifting Data Streams. 
Machine Learning and Knowledge Discovery in Databases. ECML PKDD 2016.

\bibitem[18]{18}
W. Zuo, D. Zhang, K. Wang:
On kernel difference-weighted k-nearest neighbor classification.
Pattern Anal Applic (2008) 11: 247.

\bibitem[19]{19}
A. Temko, G. Lightbody, G. Boylan, W. Marnane:
Online EEG channel weighting for detection of seizures in the neonate.
Conf Proc IEEE Eng Med Biol Soc. 2011:1447-50.

\bibitem[20]{20}
J. Shyuu, Jhing-Fa Wang and Chung-Hsien Wu: 
A channel-weighting method for speech recognition using wavelet decompositions.
Circuits and Systems, 1994. APCCAS '94., 1994 IEEE Asia-Pacific Conference on, Taipei, 1994, pp. 519-523.
\end{thebibliography}
\end{document}